\documentclass{article}
\usepackage[numbers, compress]{natbib}  % natbib must be defined here, not in macros
\usepackage[preprint]{neurips_2023}
\usepackage[utf8]{inputenc} % allow utf-8 input
\usepackage[T1]{fontenc}    % use 8-bit T1 fonts
\usepackage{microtype}
\usepackage{graphicx}  
\usepackage{booktabs}
\usepackage{hyperref}
\usepackage[dvipsnames]{xcolor}
\usepackage{algorithm}
\usepackage{algcompatible}
\usepackage{amsmath}
\usepackage{soul}
\usepackage[authormarkuptext=name,addedmarkup=bf,authormarkupposition=left]{changes}
\usepackage{xspace}
\usepackage{dsfont}
\usepackage{bm}
\usepackage{bbm}
\usepackage{nicefrac}       % compact symbols for 1/2, etc.
\usepackage[scaled=.8]{beramono} % reduce \texttt size
\usepackage{amsthm}
\usepackage{amsfonts}
\usepackage{amssymb}
\usepackage{array}
\usepackage{mathtools}
\usepackage{caption}
\usepackage{subcaption}
\usepackage{fontawesome}
\usepackage{thmtools}
\usepackage{thm-restate}
\usepackage{lipsum}
\usepackage{pifont}
\usepackage{makecell}
\usepackage{tablefootnote}
\usepackage[tight-spacing=true]{siunitx}
\usepackage{multirow}
\usepackage{afterpage}
\usepackage{tabularx} 
\usepackage{url}
\usepackage{tikz}
\usepackage{indentfirst}
\usepackage{cleveref}
\usepackage{enumitem}
\usepackage{epsdice}
\usepackage{wrapfig}
\usepackage[export]{adjustbox}% http://ctan.org/pkg/adjustbox
\usepackage{xparse}

\makeatletter
\newcommand{\subalign}[1]{%
  \vcenter{%
    \Let@ \restore@math@cr \default@tag
    \baselineskip\fontdimen10 \scriptfont\tw@
    \advance\baselineskip\fontdimen12 \scriptfont\tw@
    \lineskip\thr@@\fontdimen8 \scriptfont\thr@@
    \lineskiplimit\lineskip
    \ialign{\hfil$\m@th\scriptstyle##$&$\m@th\scriptstyle{}##$\hfil\crcr
      #1\crcr
    }%
  }%
}
\makeatother

\Crefname{figure}{Figure}{Figures}
\crefname{figure}{Figure}{Figures}
\crefname{table}{Table}{Tables}
\crefformat{table}{Table~#1}
\crefformat{equation}{equation (#1)}
\Crefformat{Equation}{Equation (#1)}
\crefformat{algorithm}{Algorithm~#1}

\hypersetup{
  colorlinks=true,
  citecolor=Blue,
  linkcolor=Blue,
  urlcolor=Blue
}

\newcommand\sdots{\hbox to 1em{.\hss.\hss.}} % dots with smaller horizontal space
\DeclareMathAlphabet\mathbfcal{OMS}{cmsy}{b}{n} % Bold mathcal
\newcommand{\smallsum}{\textstyle\sum\limits}
\DeclareMathSymbol{\shortminus}{\mathbin}{AMSa}{"39}

\setitemize[0]{leftmargin=*}

\definecolor{mygreen}{RGB}{43, 138, 62}
\definecolor{myred}{RGB}{201, 42, 42}

\newcommand*\colourcheck[1]{%
  \expandafter\newcommand\csname #1check\endcsname{\textcolor{#1}{\ding{51}}}%
}
\colourcheck{mygreen}

\newcommand*\colourmark[1]{%
  \expandafter\newcommand\csname #1mark\endcsname{\textcolor{#1}{\ding{55}}}%
}
\colourmark{myred}

\makeatletter
\let\@fnsymbol\@arabic
\makeatother

\title{MultiPrompter: Cooperative Prompt Optimization with Multi-Agent Reinforcement Learning}

\author{
  Dong-Ki Kim\thanks{LG AI Research\;\;${}^{2}$University of Michigan--Ann Arbor}\\
  \texttt{dkkim@lgresearch.ai}
  \And
  Sungryull Sohn\footnotemark[1]\\
  \texttt{srsohn@lgresearch.ai}
  \And
  Lajanugen Logeswaran\footnotemark[1]\\
  \texttt{llajan@lgresearch.ai}
  \And
  Dongsub Shim\footnotemark[1]\\
  \texttt{dongsub.shim@lgresearch.ai}
  \And
  Honglak Lee\footnotemark[1]\textsuperscript{\:\:\:,}\footnotemark[2]\\
  \texttt{honglak@lgresearch.ai}
}

\begin{document}
\maketitle
\begin{abstract}
Recently, there has been an increasing interest in automated prompt optimization based on reinforcement learning (RL). This approach offers important advantages, such as generating interpretable prompts and being compatible with black-box foundation models. However, the substantial prompt space size poses challenges for RL-based methods, often leading to suboptimal policy convergence. This paper introduces MultiPrompter, a new framework that views prompt optimization as a cooperative game between prompters which take turns composing a prompt together. Our cooperative prompt optimization effectively reduces the problem size and helps prompters learn optimal prompts. We test our method on the text-to-image task and show its ability to generate higher-quality images than baselines. 
\end{abstract}

\section{Introduction}\label{sec:introduction}
Foundation models are now an integral part of our daily lives, finding applications across various tasks and domains \cite{openai2023gpt4,rombach2022stablediffusion,wang2023voyager}. The driving force behind their widespread adoption is prompting. Unlike fine-tuning, which involves a resource-intensive process of updating numerous model parameters for a specific task, prompting effectively guides the model’s behavior by refining initial prompts \cite{brown2020fewshot,wei2022chainofthought,liu2023prompt}. With the growing availability of black-box models, prompting has emerged as an essential tool for interacting with foundation models.

An important goal in prompting is automated prompt optimization, removing the need for laborious manual trial-and-error efforts. Reinforcement learning (RL) \cite{sutton1998rlbook} presents a promising solution for achieving this goal by discovering prompts that outperform manually created ones through sequential optimization in the prompt space \cite{deng2022rlprompt,hao2022promptist,zhang2023tempera,dong2023pace,li2023dialogue,jung2023discrete}. Notably, RL-based methods generate interpretable prompts and are compatible with black-box foundation models. These attributes provide distinct advantages over alternative approaches like soft prompts \cite{qineisner2021learning,li2021prefix,lester2021power,liu2022ptuning}, which produce less interpretable prompts and require white-box access to the models. However, despite these exciting outcomes, we note that existing work suffers from the extensive size of the prompt space. This problem size is critical as it hinders effective exploration and generally leads to suboptimal policy convergence \cite{paduraru2021challenge}.

\begin{figure}[t]
    \centering
    \includegraphics[width=0.95\textwidth]{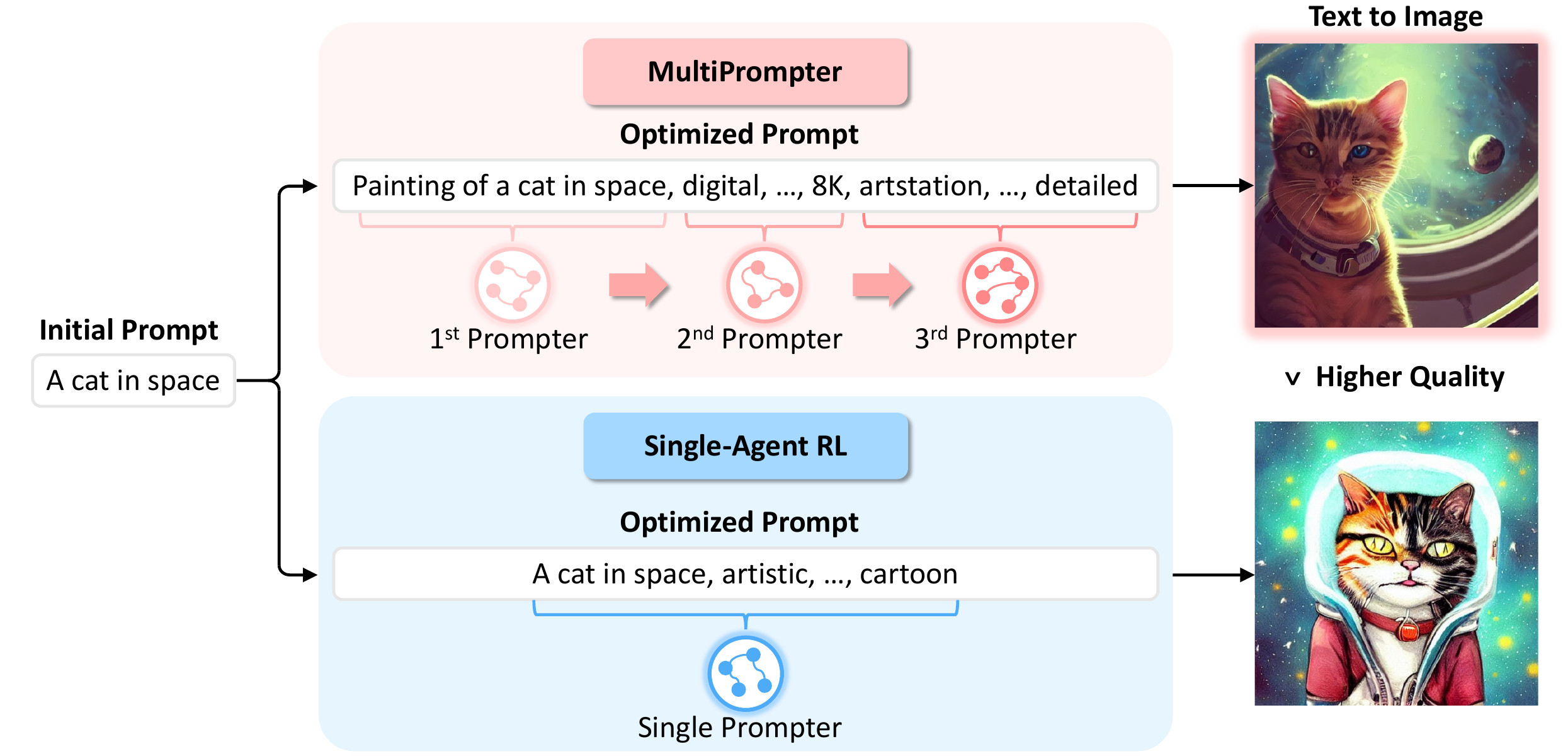}
    \vskip-0.09in
    \caption{
    In MultiPrompter, a team of prompters learns to take turns optimizing a prompt together, generating higher-quality images compared to those produced by a single-agent RL method for a text-to-image task. The images are generated using Stable Diffusion \cite{rombach2022stablediffusion}.
    }
    \label{fig:multiprompter-intro}
    \vskip-0.2in
\end{figure}

We propose a novel prompt optimization framework to address the challenge posed by the extensive prompt space. Our key idea is to view prompt optimization as a cooperative game between multiple prompters which take turns composing a prompt together (see \Cref{fig:multiprompter-intro}). In this new setting, prompters cooperatively decompose the prompt space into smaller subspaces and sequentially optimize their respective parts. As a result, our approach significantly reduces problem complexity compared to prior work that rely on a single prompter to optimize the entire prompt. To effectively learn cooperative policies within our framework, we develop a practical multi-agent RL algorithm, named \textit{MultiPrompter}. Specifically, we enable each prompter to consider the behaviors of subsequent prompters through a centralized critic \cite{lowe2017maddpg,foerster2018coma,omidshafiei2019teach,yu2022mappo} designed for cooperative prompt optimization. We show that MultiPrompter generates more optimal prompts than those produced by baselines.

\textbf{Our contribution.} In summary, this paper makes the following main contributions:
\begin{itemize}[leftmargin=*, wide, labelindent=0pt, topsep=0pt]
    \setlength\itemsep{0em}
    \item \textbf{Formulation of cooperative prompt optimization (\Cref{sec:problem-statement}).} We introduce a new cooperative game in which multiple prompters work as a team to enhance a prompt by sequentially optimizing it. This cooperative approach effectively reduces problem complexity in contrast to a single-agent RL approach and facilitates the process of finding optimal prompts. 
    \item \textbf{Algorithm for learning cooperative prompt optimization (\Cref{sec:method}).} Learning multiple prompters in our setting requires each prompter to take into account the behaviors of others. Otherwise, undesirable outcomes can arise as prompters may greedily optimize a prompt. To address this, MultiPrompter develops an actor-critic framework with a centralized critic that considers the next prompter's actions to predict the value accurately. 
    \item \textbf{Comprehensive evaluation of MultiPrompter (\Cref{sec:evaluation}).} Our results highlight that MultiPrompter outperforms a single-agent RL baseline in the text-to-image generation task, obtaining higher rewards and the ability to optimize longer prompts. Additionally, we consider a variant of MultiPrompter, which applies a competitive game between prompters, and show that this competitive prompt optimization is less effective than our cooperative formulation.
\end{itemize}

%%%%%%%%%%%%%%%%%%%%%%%%%%%%%%%%%%%%%%%%%%%%%%%%%%%%%%%%%
%%%%%%%%%%%%%%%%%%%%%%%%%%%%%%%%%%%%%%%%%%%%%%%%%%%%%%%%%

\section{Problem Statement: Cooperative Prompt Optimization Game}\label{sec:problem-statement}
\textbf{Overview.} We introduce the new concept of cooperative prompt optimization: given an initial prompt $\bm{x}$, a team of $n$ prompters generates an optimized prompt that consists of multiple subprompts $\bm{y}\!=\!(\bm{\tilde{y}^1},...,\bm{\tilde{y}^n})$ (see \Cref{fig:multiprompter-intro}). Specifically, the first prompter optimizes the initial parts of a prompt and passes the baton to the next prompter as needed. Then, the second prompter continues the optimization from where the previous prompter left off. This process repeats until the last prompter finishes its turn or the length of an optimized prompt exceeds a token limit. The team's objective is to generate an optimized prompt such that it achieves a high score according to a performance metric. 

\textbf{Definition.} We formally define a cooperative prompt optimization game between $n$ prompters as a tuple $\mathcal{G}_n\!=\!\langle\mathbfcal{I},\mathcal{S},\mathcal{V},\mathcal{T},\mathcal{R}\rangle$;
$\mathbfcal{I}\!=\!(1,...,n)$ is the set of $n$ prompters;
$\mathcal{S}$ is the prompt space;
$\mathcal{V}$ is the vocabulary;
$\mathcal{T}\!:\!\mathcal{S}\!\times\!\mathcal{V}\!\mapsto\!\mathcal{S}$ is the deterministic prompt transition function; and
$\mathcal{R}$ is the reward function. 
We also define an index $i\!\in\!\mathbfcal{I}$ that points to the active prompter which is currently taking its turn. At the beginning of the game, prompters are given an initial prompt $\bm{x}\!=\!(x_1,...,x_K)$ and the index $i$ is reset to one. At each timestep $t$, the $i$-th prompter decides a discrete token $y_t\!\in\!\mathcal{V}$ according to its stochastic policy $y_t\!\sim\!\pi^i(\cdot|\bm{x},\bm{y_{1:t\shortminus 1}}; \theta^i)$ parameterized by $\theta^i$, where $\bm{y_{1:t\shortminus 1}}\!=\!(y_1,...,y_{t \shortminus 1})$. An action $y_t$ then yields the prompt transition from ($\bm{x},\bm{y_{1:t\shortminus 1}}$) to ($\bm{x},\bm{y_{1:t}}$). If $y_t$ corresponds to an end-of-sequence token, then $i$-th prompter finishes generating its subprompt $\bm{\tilde{y}^{i}}$. The index $i$ is also updated to pass the turn to the next prompter (i.e., $i\!\leftarrow\!i\!+\!1$). The game ends when all prompters finish their turns or the size of an optimized prompt $\bm{y}$ is over the token limit $T$. The team shares a reward according to $\mathcal{R}(\bm{x},\bm{y})$, which measures the quality of an optimized prompt $\bm{y}$ at the end of the game.

\textbf{Benefit of problem size reduction.} The natural outcome of our cooperative game is the generation of a decomposed optimized prompt, which consists of subprompts: $\bm{y}\!=\!(\bm{\tilde{y}^1},...,\bm{\tilde{y}^n})$. Thanks to this cooperative prompt decomposition, each prompter $i$ in MultiPrompter simply has to search for the desired tokens within its subprompt $\bm{\tilde{y}^{i}}$, which contains fewer tokens than the full prompt $\bm{y}$. As a result, we note that our formulation substantially reduces problem complexity compared to optimizing a prompt using single-agent RL methods \cite{deng2022rlprompt,abdulhai2022cradol}:
\begin{align}
\begin{split}
    \underbrace{\smallsum_{i\in\mathbfcal{I}} |\mathcal{V}|^{|\bm{\tilde{y}^i}|}}_{\substack{\text{Multi prompter} \\ \text{problem size}}}\;\;\;\ll \underbrace{\vphantom{\smallsum_{i\in\mathbfcal{I}}}|\mathcal{V}|^{|\bm{y}|}}_{\substack{\text{Single prompter} \\ \text{problem size}}}
\end{split}
\end{align}
where $|\!\cdot\!|$ denotes the size of a set. In the following section, we leverage this inherent benefit of our cooperative prompt optimization game and develop an algorithm for learning cooperative policies.

\section{MultiPrompter: Learning Cooperative Prompt Optimization Policies}\label{sec:method}
This section introduces our multi-agent learning algorithm, named MultiPrompter, designed to learn cooperative policies that decompose and optimize a prompt together. We first outline each prompter's objective in our cooperative prompt optimization game. We then detail our centralized critic that enables each prompter to consider the actions and learning of others, thus facilitating successful cooperation. We provide additional details, including pseudocode, in \Cref{appendix:multiprompter-detail}.

\textbf{MultiPrompter objective.} The objective of each prompter's policy $\pi^i$ is to find policy parameters $\theta^{i}$ that maximize the expected return:
\begin{gather}
\max_{\theta^{i}}\mathbb{E}_{\bm{x},\bm{y}\sim p(\cdot|\bm{\theta})}\big[\mathcal{R}(\bm{x},\bm{y})\big], \text{ where }
p(\bm{x},\bm{y}|\bm{\theta})=p(\bm{x})\prod\nolimits_{i=1}^{n}\prod\nolimits_{t=t^{i}_{\text{bos}}}^{t^{i}_{\text{eos}}}\pi^{i}(y_{t}|\bm{x},\bm{y_{1:t\shortminus 1}};\theta^{i}),\label{eqn:objective}
\end{gather}
where $t^{i}_{\text{bos}}$ and $t^{i}_{\text{eos}}$ denote the beginning and end timesteps of $\bm{\tilde{y}^{i}}$, respectively.
Leveraging REINFORCE \cite{williams1992reinforce}, we derive a policy gradient with respect to the objective in \Cref{eqn:objective}:
\begin{align}\label{eqn:policy-gradient}
\begin{split}
\nabla_{\theta^{i}}\mathbb{E}_{\bm{x},\bm{y}\sim p(\cdot|\bm{\theta})}\big[\smallsum\nolimits_{t=t^{i}_{\text{bos}}}^{t^{i}_{\text{eos}}}\log\pi^{i}(y_{t}|\bm{x},\bm{y_{1:t\shortminus 1}};\theta^{i})\mathcal{A}^{i}_{t}(\bm{x},\bm{y})\big],
\end{split}
\end{align}
where $\mathcal{A}^{i}_{t}(\bm{x},\bm{y})$ denotes the advantage function, which we will detail our choice in the next paragraph. 

\setlength{\columnsep}{8pt}
\begin{wrapfigure}[13]{r}{0.45\textwidth}
  \vskip-0.15in
  \centering
  \includegraphics[width=\linewidth]{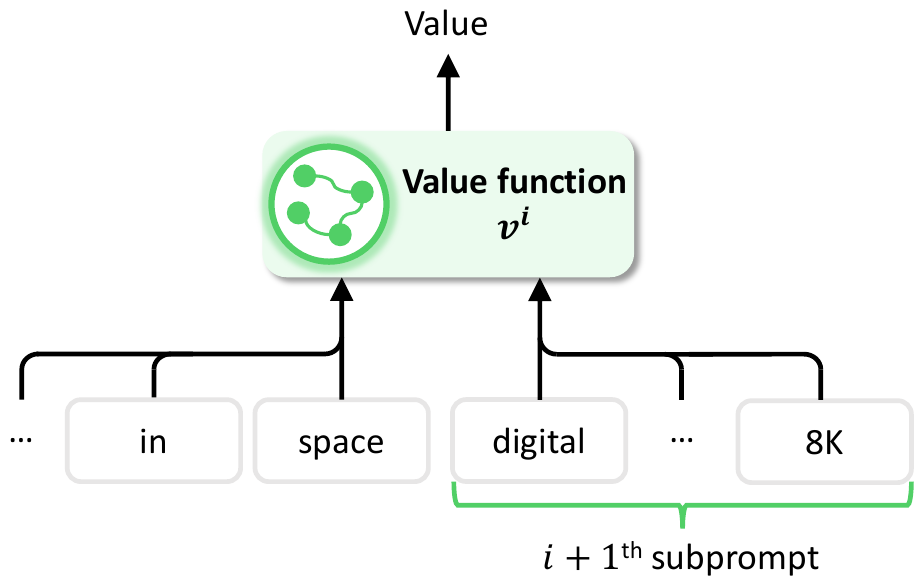}
  \vskip-0.05in
  \caption{Our value function additionally considers a subprompt of the next prompter $\bm{\tilde{y}^{i\!+\!1}}$.}
  \label{fig:centralized critic}
\end{wrapfigure}
\textbf{Centralized critic.} The generalized advantage estimation \cite{schulman2018gae} shows that the advantage function can be effectively estimated with low variance using a value function. A single-agent RL approach uses a value function $v(\bm{x},\bm{y_{1:t\shortminus 1}};\phi)$ parameterized by $\phi$ \cite{hao2022promptist}, but this form neglects the presence of other prompters, even though the reward is jointly affected by all of them. To consider policies and learning of subsequent prompters, MultiPrompter uses a value function $v^{i}(\bm{x},\bm{y_{1:t\shortminus 1}},\bm{\tilde{y}^{i+1}};\phi^{i})$ that enables a prompter $i$ to consider a prompter $i+1$'s policy (see \cref{fig:centralized critic}). We have a design choice regarding the number of next prompters to consider in the value function, and we empirically find that including the next prompter's information results in effective training. Note that the centralized value function is only utilized during training, which takes additional information to accurately predict the value. Each prompter's policy remains decentralized, so MultiPrompter follows the centralized training with decentralized execution structure \cite{lowe2017maddpg,foerster2018coma,omidshafiei2019teach,yu2022mappo}. Lastly, we provide value function optimization and other related details in \Cref{appendix:multiprompter-detail}. 

\section{Evaluation}\label{sec:evaluation}
\textbf{Experiment setup.} Following the experimental settings by \cite{hao2022promptist}, we use a reward function $\mathcal{R}(\bm{x},\bm{y})$ that consists of the relevance score (i.e., measures the degree of relevance between an initial prompt $\bm{x}$ and an image generated $\bm{y}$) and the aesthetic score (i.e., measures aesthetical preference of an image generated by $\bm{y}$ over another image generated by $\bm{x}$). We implement both policies and value functions using GPT-2 \cite{radford2019gpt2} and initialize them from the weights fine-tuned with manually engineered prompts \cite{hao2022promptist}. We compare methods using the COCO dataset \cite{lin2014cocodataset}. We refer to \Cref{appendix:experiment-detail} for more details.
\textbf{Baseline.} We compare MultiPrompter with the following baselines:
\begin{itemize}[leftmargin=*, wide, labelindent=0pt, topsep=0pt]
    \setlength\itemsep{0em}
    \item \textbf{Manual prompt \cite{hao2022promptist}.} A fine-tuned method with human-engineered prompts from Lexica \cite{lexica}.
    \item \textbf{Promptist \cite{hao2022promptist}.} A single-agent RL method that trains a GPT-2 prompter based on PPO \cite{schulman2017ppo}.
    \item \textbf{Competition.} Our variant of MultiPrompter that applies competition between prompters \cite{bansal2018emergent}.
\end{itemize}
We omit soft prompt methods \cite{qineisner2021learning,li2021prefix,lester2021power,liu2022ptuning} in our evaluation, because our work focuses on generating interpretable prompts without access to foundation models. 

%%%%%%%%%%%%%%%%%%%%%%%%%%%%%%%%%%%%%%%%%%%%%%%%
\begin{figure}[t]
\begin{minipage}[t]{0.56\textwidth}
\vskip-1.35in
\begin{table}[H]
    \centering
    \renewrobustcmd{\bfseries}{\fontseries{b}\selectfont}
    \sisetup{detect-weight,mode=text,group-minimum-digits=4}
    {\small
	\tabcolsep=0.10cm
	\begin{tabular}[t]{l
		S[separate-uncertainty,table-figures-uncertainty=1,table-format=2.2(2)]
		S[separate-uncertainty,table-figures-uncertainty=1,table-format=3(3)]
		S[separate-uncertainty,table-figures-uncertainty=1,table-format=2.2(2)]
		}
		\toprule
		Algorithm & \text{Multi-agent?} & \text{Collaborative?} & \text{Test reward} \\ \midrule
		Manual prompt &  \myredmark &  \myredmark & $\shortminus$ 0.68 \pm 0.06 \\
		Promptist &  \myredmark &  \myredmark & 0.28 \pm 0.11 \\
		Competition &  \mygreencheck &  \myredmark & 0.36 \pm 0.12 \\
		\midrule
		MultiPrompter    &   \mygreencheck &  \mygreencheck & \bfseries 0.76 \pm 0.10 \\
		\bottomrule
	\end{tabular}
	}
    \vskip 0.32in
    \caption{Test performance across various methods. MultiPrompter achieves a statistically significant performance improvement through cooperative prompt optimization.}
    \label{table:results} 
\end{table}
\end{minipage}
\hfill
\begin{minipage}[t]{0.42\textwidth}
\centering
\includegraphics[width=0.8\textwidth]{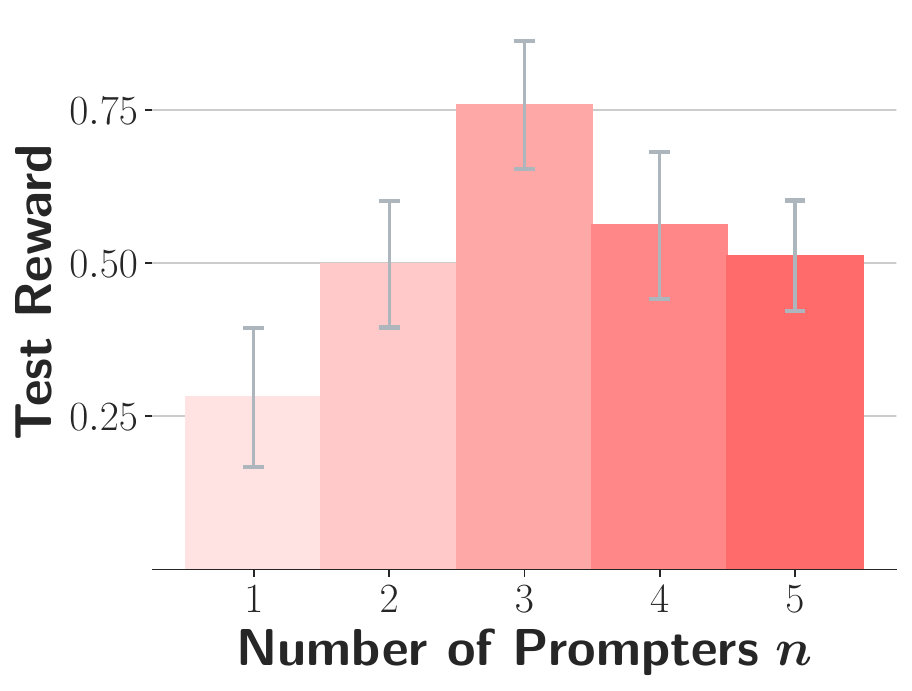}
\vskip-0.03in
\caption{MultiPrompter's performance w.r.t. $n$ shows a trade-off between problem size reduction and cooperation complexity.}
\label{fig:result-n-prompter}
\end{minipage}
\vskip-0.33in
\end{figure}
%%%%%%%%%%%%%%%%%%%%%%%%%%%%%%%%%%%%%%%%%%%%%%%%

\textbf{Question 1.} \textit{How effective is MultiPrompter compared to a single-agent RL baseline?}

\Cref{table:results} provides a summary of the test time performance. For the training performance across multiple seeds, we refer to \Cref{fig:train-performance} in the Appendix. Our main observation is that MultiPrompter outperforms both the single-agent RL and manual prompt baselines. To understand how MultiPrompter generates more optimal prompts compared to Promptist, we examine the number of optimized tokens by each method. While Promptist converges to optimizing an average of $57.01$ tokens, MultiPrompter optimizes an average of $69.46$ tokens (refer to \Cref{fig:example} in the Appendix for an example). This result indicates that the single-agent RL approach generally suffers from the extensive prompt space, converging to a suboptimal policy that adds only a limited number of modifiers to the original prompt. In contrast, MultiPrompter successfully overcomes this challenge through cooperative prompt optimization and finds a greater number of effective modifiers compared to Promptist.

\textbf{Question 2.} \textit{What about posing prompt optimization as a competitive game?}

When considering prompt optimization from a multi-agent perspective, cooperative optimization is not the only approach, but there is also the alternative of competitive prompt optimization. We consider a competitive setting in which each prompter optimizes a full prompt individually and then compares its optimized prompt to the output of another competing prompter (refer to \Cref{appendix:self-play-detail} for details). As \Cref{table:results} shows, we find that competitive prompt optimization produces a slight improvement over the single-agent RL method but it is not as effective as cooperative prompt optimization, primarily due to its lack of the ability to decompose the prompt space.

\textbf{Question 3.} \textit{How does MultiPrompter's performance change with respect to the number of prompters?}

\Cref{fig:result-n-prompter} shows an analysis of MultiPrompter in relation to the number of prompters $n$. There are two notable observations. First, we observe a trend in which test performance increases with $n$, but then decreases after $n\!=\!3$. This result suggests a general trade-off in cooperative prompt optimization: while the prompt space size reduces as $n$ increases (as discussed in \Cref{sec:problem-statement}), the complexity of learning cooperation between prompters increases. Our future work includes incorporating recent advances in multi-agent RL \cite{yang2018mean,rabinowitz2018tom,kim2021metamapg,kim22further} to effectively address the learning complexity with increasing $n$. Second, we note that MultiPrompter with any $n\!>\!1$ performs better than a single-agent RL approach (i.e., $n\!=\!1$).

\textbf{Question 4.} \textit{How important is it to learn the prompt decomposition and have the centralized critic?}

\begin{wraptable}[12]{r}{0.55\textwidth}
    \vskip-0.15in
    \centering
    \renewrobustcmd{\bfseries}{\fontseries{b}\selectfont}
    \sisetup{detect-weight,mode=text,group-minimum-digits=4}
    {\small
	\tabcolsep=0.10cm
	\begin{tabular}[t]{
		S[separate-uncertainty,table-figures-uncertainty=1,table-format=2.2(2)]
		S[separate-uncertainty,table-figures-uncertainty=1,table-format=3(3)]
		S[separate-uncertainty,table-figures-uncertainty=1,table-format=2.2(2)]
		}
		\toprule
		\text{Learned decomposition?} & \text{Centralized critic?} & \text{Test reward} \\ \midrule
		\myredmark &  \myredmark & 0.40 \pm 0.12 \\
		\mygreencheck &  \myredmark & 0.39 \pm 0.12 \\
		\myredmark &  \mygreencheck & 0.59 \pm 0.16 \\
		\midrule
		\mygreencheck &  \mygreencheck & \bfseries 0.76 \pm 0.10 \\
		\bottomrule
	\end{tabular}
	}
    \caption{Ablation study of MultiPrompter. By learning to collaboratively decompose the prompt space and employing the centralized critic to take into account the behaviors of the next prompter, MultiPrompter achieves the best performance.}
    \label{table:ablation} 
\end{wraptable}

MultiPrompter achieves cooperative prompt optimization by learning the dynamic decomposition of the prompt space. This process involves each prompter learning the appropriate moment to finish its turn and pass the baton to the next prompter, while taking into account the actions of the following prompter through the centralized critic. \Cref{table:ablation} presents an ablation analysis of MultiPrompter, examining its performance in learning a flexible prompt space decomposition versus manual decomposition (i.e., fixing each prompter's token limit as $T/n$). We also study the impact of using the centralized critic in learning policies. Two notable observations emerge. First, without the centralized critic, each prompter cannot consider the behaviors of other prompters, resulting in ineffective collaboration. Second, while utilizing the centralized critic improves performance, the combination of this approach with the dynamic prompt space decomposition as in MultiPrompter leads to the best performance.

\section{Conclusion}
In this paper, we have introduced MultiPrompter to address the extensive prompt space size in RL-based prompt optimization. The key idea is to learn multiple cooperative prompters that optimize a prompt together. We tested our method on various settings and showed that MultiPrompter consistently outperforms baseline approaches in the text-to-image domain.

%%%%%%%%%%%%%%%%%%%%%%%%%%%%%%%%%%%%%%%%%%%%%%%%%%%%%%%%%%%%
%% REFERENCE
\bibliographystyle{unsrt}
\bibliography{main}

\newpage
\appendix
\section{Additional MultiPrompter Details}\label{appendix:multiprompter-detail}
\subsection{Optimization}
The advantage function $\mathcal{A}^{i}_{t}(\bm{x},\bm{y})$ in \Cref{eqn:policy-gradient} can be effectively estimated using the generalized advantage estimation \cite{schulman2018gae} using a value function:
\begin{gather}
\mathcal{A}^{i}_{t}(\bm{x},\bm{y})=\smallsum\nolimits_{l=0}^{|\bm{\tilde{y}^{i}}|-1}\lambda^{l}\delta^{i}_{t+l}\\
\delta^{i}_{t+l}=
\begin{cases}
    \mathcal{R}(\bm{x},\bm{y})\!-\!v^{i}(\bm{x},\bm{y_{1:t+l-1}},\bm{\tilde{y}^{i+1}};\phi^{i})\hspace{2.23cm}\text{if } t+l\!=\!t^{i}_{\text{eos}},\\
    v^{i}(\bm{x},\bm{y_{1:t+l}},\bm{\tilde{y}^{i+1}};\phi^{i})\!-\!v^{i}(\bm{x},\bm{y_{1:t+l-1}},\bm{\tilde{y}^{i+1}};\phi^{i}) \;\text{else,}
\end{cases}
\end{gather}
where we assume the discount factor $\gamma\!=\!1$ and sparse reward function $\mathcal{R}(\bm{x},\bm{y})$. We update value function parameters by minimizing the standard squared-error loss with the target value $v_{\text{target},t}$:
\begin{align}\label{eqn:value-loss}
\begin{split}
\mathcal{L}_{v}(\phi^{i})=\mathbb{E}_{\bm{x},\bm{y}\sim p(\cdot|\bm{\theta})}\big[\smallsum\nolimits_{t=t^{i}_{\text{bos}}}^{t^{i}_{\text{eos}}}(v(\bm{x},\bm{y_{1:t-1}},\bm{y^{i+1}};\phi^{i})-v_{\text{target},t})^{2}\big].
\end{split}
\end{align}
In this work, we apply PPO \cite{schulman2017ppo} to update policies and value functions with respect to \Cref{eqn:policy-gradient} and \Cref{eqn:value-loss}, respectively.

\subsection{Reward Engineering}\label{sec:reward-engineering}
Since MultiPrompter is a multi-agent learning approach, our work is also affected by the credit assignment issue in multi-agent RL \cite{foerster2018coma,chang2003credit,sunehag2018value}, where certain agents do not actively participate in cooperation. In particular, we observe that a prompter may optimize most or the entire prompt by itself, thereby not providing opportunities to subsequent teammates. To effectively address this issue, we add the following cooperation reward, which computes the entropy with respect to the lengths of the subprompts, to the original reward:
\begin{align}\label{eqn:entropy-reward}
\begin{split}
\mathcal{R}_{\text{cooperation}}(\bm{y})=\mathcal{H}(\bm{y})=-\smallsum_{i\in\mathcal{I}}\big((|\bm{\tilde{y}^{i}}|/|\bm{y}|)\log(|\bm{\tilde{y}^{i}}|/|\bm{y}|)\big)/\log n.
\end{split}
\end{align}
Intuitively, this reward function encourages prompters to evenly decompose the prompt space such that they collectively optimize a prompt.

\subsection{Pseudocode} 
\begin{algorithm}[H]
    \caption{MultiPrompter}\label{alg:algorithm}  
    \small
    \begin{algorithmic}[1]
    \REQUIRE policy parameters $\bm{\theta}=(\theta^1,...,\theta^n)$, value parameters $\bm{\phi}=(\phi^1,...,\phi^n)$, token limit $T$
    \WHILE{not converged}
	\STATE \textit{\# perform episode reset}
        \STATE get an initial prompt $\bm{x}\sim p(\bm{x})$
        \STATE reset index $i=1$ and timestep $t=1$
	\STATE \textit{\# start prompt optimization}
	\WHILE{$i\leq n$}
	    \STATE \textit{\# select current token $y_t$}
		\IF{$t\leq T$}
		    \STATE sample current token from an active prompter $y_{t}\sim\pi^i(\cdot|\bm{x},\bm{y_{1:t-1}};\theta^{i})$ 
            \ELSE
		    \STATE set current token as an EOS token $y_{t}=y_{\text{eos}}$ 
            \ENDIF
	    \STATE \textit{\# update timestep $t$ and index $i$}
            \STATE update timestep $t\leftarrow t+1$
		\IF{$y_{t}$ corresponds to an EOS token $y_{\text{eos}}$}
                \STATE update index $i\leftarrow i+1$
            \ENDIF
	\ENDWHILE
	\STATE compute a team reward $\mathcal{R}(\bm{x},\bm{y})$
	\STATE \textit{\# train prompters}
        \FOR{$i=1,...,n$}
	    \STATE train policy parameters $\theta^i$ according to \Cref{eqn:policy-gradient}
	    \STATE train value function parameters $\phi^i$ according to \Cref{eqn:value-loss}
        \ENDFOR
    \ENDWHILE
\end{algorithmic}
\end{algorithm}

%%%%%%%%%%%%%%%%%%%%%%%%%%%%%%%%%%%%%%%%%%%%%%%%%%%%%%%%%
\begin{figure}[t]
    \centering
    \includegraphics[width=0.95\textwidth]{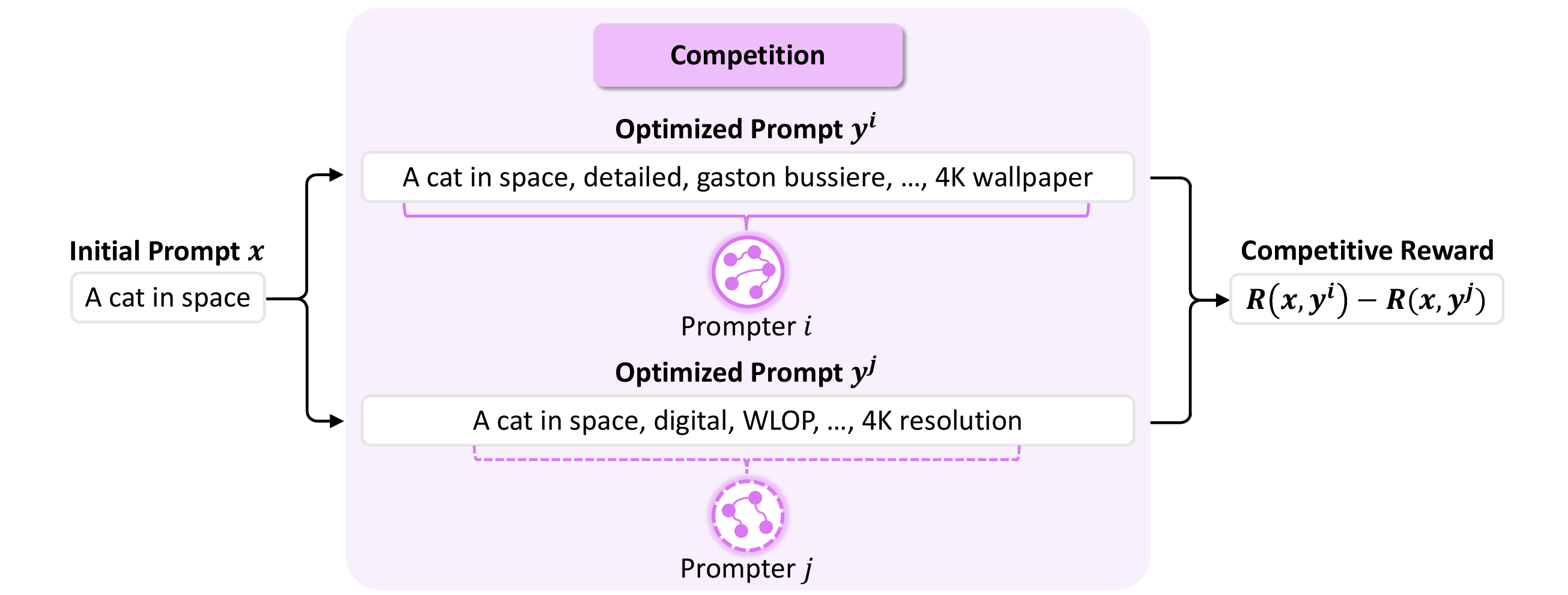}
    % \vskip-0.095in
    \caption{
    We present a competitive prompt optimization approach, where a prompter $i$ competes against another prompter $j$ by comparing their respective optimized prompts.
    }
    \label{fig:multiprompter-selfplay}
    \vskip-0.2in
\end{figure}
%%%%%%%%%%%%%%%%%%%%%%%%%%%%%%%%%%%%%%%%%%%%%%%%%%%%%%%%%

\section{Additional Evaluation Details}\label{appendix:experiment-detail}
\subsection{Competition Baseline Details}\label{appendix:self-play-detail}
In this work, we contrast our cooperative prompt optimization approach with an alternative setting of competitive prompt optimization. Specifically, we design a competitive setting, where a prompter $i$ generates its own full prompt $\bm{y^i}$ and then compares its response against another prompter $j$'s optimized prompt $\bm{y^j}$. A prompter $i$ receives a competitive reward, defined as $\mathcal{R}(\bm{x},\bm{y^i})-\mathcal{R}(\bm{x},\bm{y^j})$, as a result of the comparison (see \Cref{fig:multiprompter-selfplay}). We apply the self-play technique \cite{bansal2018emergent,baker2020hideseek} to train a prompter that competes against its former copies and comparable skill levels. We also use the centralized critic with a form $v^{i}(\bm{x},\bm{y_{1:t\shortminus 1}},\bm{y^{j}};\phi^{i})$ to enable a prompter $i$ to consider another competition prompter $j$'s policy.

\subsection{Reward and Hyperparameter Details}
\textbf{Reward details.} We follow \cite{hao2022promptist} and use a reward function that measures the quality of an optimized prompt $\bm{y}$. Specifically, the reward function $\mathcal{R}(\bm{x},\bm{y})$ first computes the relevance score:
\begin{align}\label{eqn:relevance-score}
\begin{split}
\mathcal{R}_{\text{relevance}}(\bm{x},\bm{y})=\mathbb{E}_{img_{\bm{y}} \sim\mathcal{M}(\bm{y})}\big[\min(20 f_{\text{CLIP}}(\bm{x},img_{\bm{y}})-5.6,0)\big],
\end{split}
\end{align}
where $img_{\bm{y}}$ refers to an image generated according to $\bm{y}$, $\mathcal{M}$ refers to a text-to-image model (e.g., Stable Diffusion \cite{rombach2022stablediffusion}), and $f_{\text{CLIP}}$ refers to the CLIP similarity function \cite{radford2021clip}. The relevance score measures the degree of relevance between $img_{\bm{y}}$ and the initial prompt $\bm{x}$. The reward function also computes the aesthetic score that measures the aesthetical preference of $img_{\bm{y}}$ over $img_{\bm{x}}$:
\begin{align}\label{eqn:aesthetic-score}
\begin{split}
\mathcal{R}_{\text{aesthetic}}(\bm{x},\bm{y})=\mathbb{E}_{img_{\bm{x}} \sim\mathcal{M}(\bm{x}),img_{\bm{y}} \sim\mathcal{M}(\bm{y})}\big[f_{\text{aesthetic}}(img_{\bm{y}})-f_{\text{aesthetic}}(img_{\bm{x}})\big],
\end{split}
\end{align}
where $f_{\text{aesthetic}}$ refers to the aesthetic predictor \cite{laion}.
Finally, the reward function sums the two scores: $\mathcal{R}(\bm{x},\bm{y})\!=\!\mathcal{R}_{\text{relevance}}(\bm{x},\bm{y})\!+\!\mathcal{R}_{\text{aesthetic}}(\bm{x},\bm{y})$. For the case of cooperative prompt optimization, we also add the cooperation reward in \Cref{sec:reward-engineering} during training: $\mathcal{R}(\bm{x},\bm{y})\!=\!\mathcal{R}_{\text{relevance}}(\bm{x},\bm{y})\!+\!\mathcal{R}_{\text{aesthetic}}(\bm{x},\bm{y})\!+\!\alpha\mathcal{R}_{\text{cooperation}}(\bm{y})$, where $\alpha$ denotes a weight.

\textbf{Hyperparameter details.}
We report important hyperparameter values in our experiments:
\begin{table}[H]
\centering
\begin{tabular}{l|l}
\textbf{Hyperparameter} & \textbf{Value} \\ \hline
Number of prompters $n$ & 1,2,3,4,5 \\
Batch size & 256 \\
Minibatch size & 128 \\
Stable Diffusion inference step & 20 \\
Token limit $T$ & 80 \\
Learning rate & 0.00001 \\
Entropy weight & 0.001\\
Discount factor $\gamma$ & 1.0 \\
GAE $\lambda$ & 0.95 \\
Cooperation reward weight $\alpha$ & 0.25 \\
\end{tabular}
\caption{Hyperparameter values used in our experiments.}
\end{table}

\subsection{Additional Result}
\textbf{Train performance.}
%%%%%%%%%%%%%%%%%%%%%%%%%%%%%%%%%%%%%%%%%%%%%%%%%%%%%%%%%
\begin{figure}[H]
    \centering
    \includegraphics[width=0.7\textwidth]{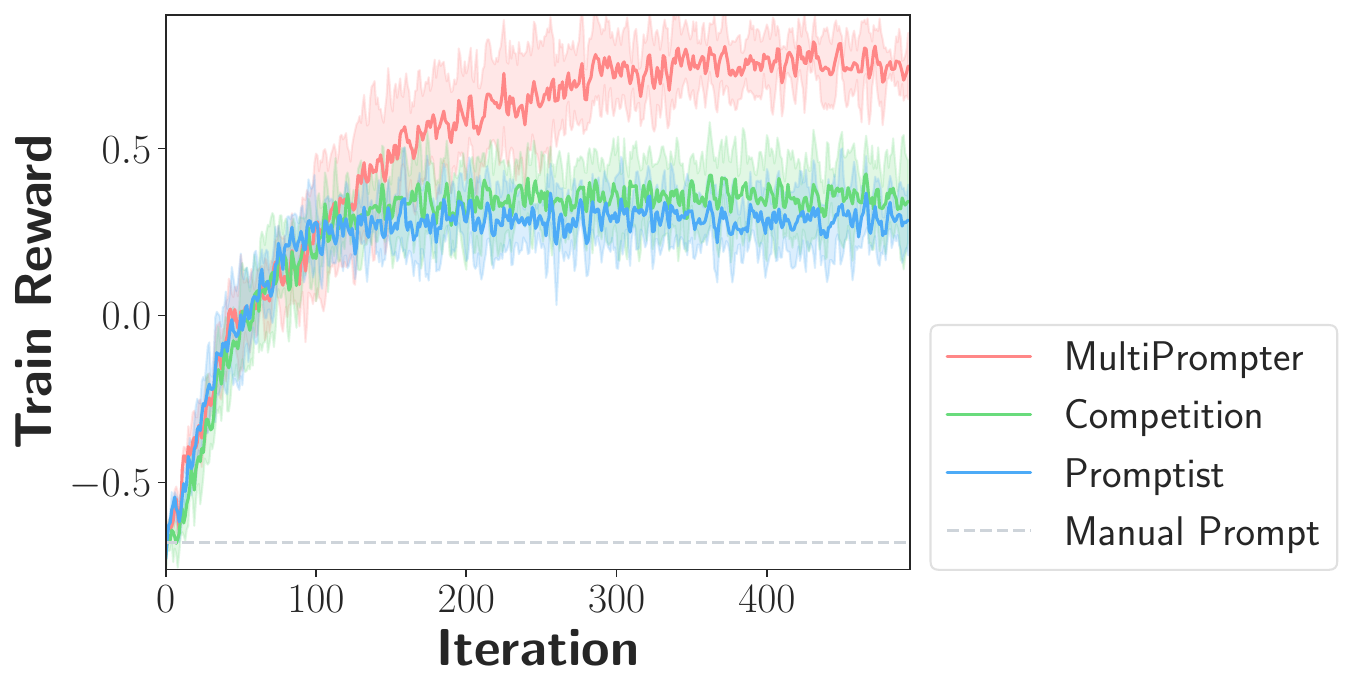}
    \caption{
    Training performance for each method. Thanks to our cooperative prompt optimization, MultiPrompter converges to a more optimal policy. The mean and standard deviation computed for 3 seeds are shown in the figure.
    }
    \label{fig:train-performance}
\end{figure}
%%%%%%%%%%%%%%%%%%%%%%%%%%%%%%%%%%%%%%%%%%%%%%%%%%%%%%%%%

\textbf{Test example.}
%%%%%%%%%%%%%%%%%%%%%%%%%%%%%%%%%%%%%%%%%%%%%%%%%%%%%%%%%
\begin{figure}[H]
    \centering
    \includegraphics[width=\textwidth]{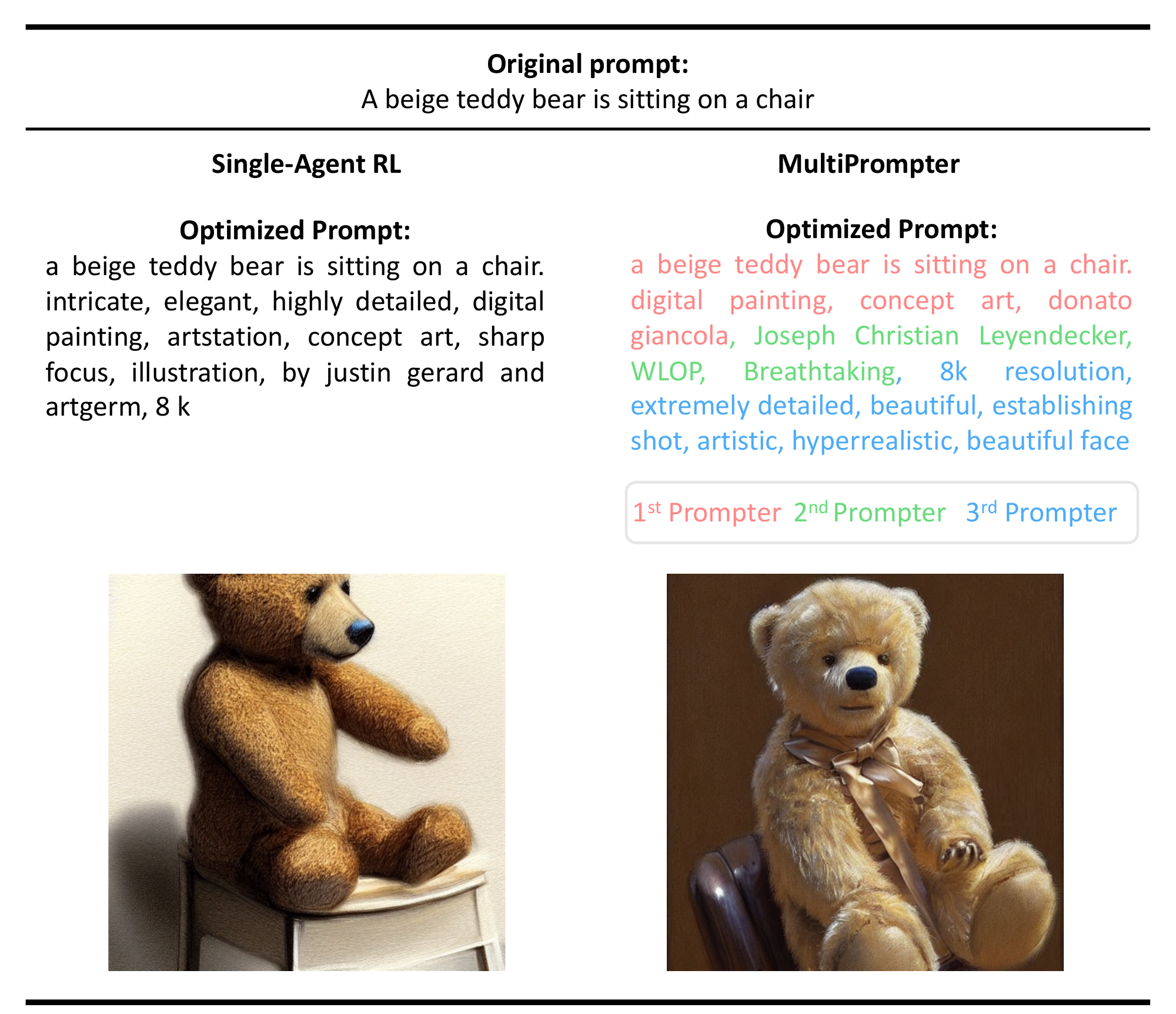}
    \caption{
    An example of an optimized prompt generated by a single-agent RL baseline and MultiPrompter. This example highlights that MultiPrompter adds a greater number of effective modifiers compared to the baseline. These images are generated using Stable Diffusion \cite{rombach2022stablediffusion}.
    }
    \label{fig:example}
\end{figure}
%%%%%%%%%%%%%%%%%%%%%%%%%%%%%%%%%%%%%%%%%%%%%%%%%%%%%%%%%
\end{document}